\documentclass[10pt,twocolumn,letterpaper]{article}

\usepackage{iccv}
\usepackage{epsfig}
\usepackage{graphicx}
\usepackage{amsmath}
\usepackage{amssymb}

\usepackage{graphicx}
\usepackage{amsmath}
\usepackage{amssymb}
\usepackage{booktabs,eucal}

\usepackage{ upgreek } %math equations

\usepackage[accsupp]{axessibility} % Improves PDF readability for those with disabilities.

\usepackage{multirow} %for tables

\def\Loss{{$\ell_1$}}

\usepackage[dvipsnames,table,xcdraw]{xcolor}
\usepackage[pagebackref=true,breaklinks=true,letterpaper=true,colorlinks,bookmarks=false]{hyperref}
\usepackage{caption}
\usepackage[capitalize]{cleveref}
\crefname{section}{Sec.}{Secs.}
\Crefname{section}{Section}{Sections}
\Crefname{table}{Table}{Tables}
\crefname{table}{Table}{Tables}

\usepackage[breaklinks=true,bookmarks=false]{hyperref}

\iccvfinalcopy % *** Uncomment this line for the final submission

\def\iccvPaperID{****} % *** Enter the ICCV Paper ID here

\ificcvfinal\fi

\begin{document}

%%%%%%%%% TITLE
\title{Robust Geometry-Preserving Depth Estimation 
Using Differentiable Rendering}

\def\SP{~~~}

\author{
Chi Zhang$^{1 \ast}$,
\SP
Wei Yin$^2$\thanks{Equal contributions.}, 
\SP 
Gang Yu$^1$\thanks{Corresponding author.},
\SP
Zhibin Wang$^1$,
\SP 
Tao Chen$^3$, \\
\SP
Bin Fu$^{1}$,
\SP
Joey Tianyi Zhou$^{5}$,
\SP
Chunhua Shen$^{4}$
\\[0.1125cm]
\normalsize 
$ ^1$ Tencent
\SP ~~~
$ ^2 $ DJI Technology
\SP ~~~
$ ^3 $ Fudan University
\normalsize 
\SP ~~~
$ ^4 $ Zhejiang University \\
\normalsize 
\SP ~~~
$ ^5 $ Centre for Frontier AI Research, A$^*$STAR 
\SP ~~~
$ ^5 $ Institute of High Performance Computing , A$^*$STAR 
\\
\\
$^1$\normalsize \tt  \{johnczhang, skicyyu, billzbwang, brianfu\}@tencent.com \\
$^2$\normalsize \tt  yvanwy@outlook.com
$^3$\normalsize \tt eetchen@fudan.edu.cn
$^4$\normalsize \tt  chunhua@me.com
$^5$\normalsize \tt  joey.tianyi.zhou@gmail.com
}

\makeatletter
\let\@oldmaketitle\@maketitle%
\renewcommand{\@maketitle}{\@oldmaketitle%
 \centering
    \includegraphics[width=1\textwidth]{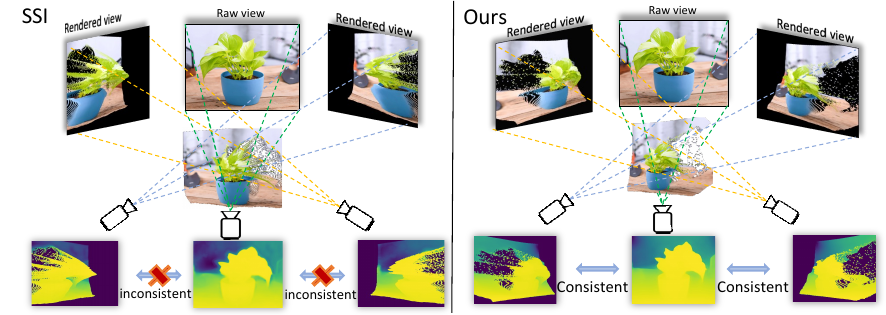}
  %  \vskip -0.5em
     \captionof{figure}{\textbf{An illustration of our motivation.}
	Zero-shot depth estimation models trained with scale-and-shift invariant (SSI) loss~\cite{ssi} produce geometry-incomplete depth prediction, which results in distorted 3D structures (left).
	A good 3D model should look realistic  across different views, such that the depth estimation results of them are consistent. Based on this intuition, 	we render novel views of the generated 3D structure and design loss functions to promote the consistency of  depth estimation across different views, which produces realistic 3D structures (right).}
    \label{fig:teaser}
    \bigskip}                   %
\makeatother

\maketitle
\ificcvfinal\thispagestyle{empty}\fi

%%%%%%%%% ABSTRACT
%%%%%%%%% ABSTRACT

\begin{abstract}
In this study, we address the challenge of 3D scene structure recovery from monocular depth estimation. While traditional depth estimation methods leverage labeled datasets to directly predict absolute depth, recent advancements advocate for mix-dataset training, enhancing generalization across diverse scenes. However, such mixed dataset training yields depth predictions only up to an unknown scale and shift, hindering accurate 3D reconstructions. Existing solutions necessitate extra 3D datasets or geometry-complete depth annotations, constraints that limit their versatility. In this paper, we propose a learning framework that trains models to predict geometry-preserving depth without requiring extra data or annotations. To produce realistic 3D structures, we  render novel views of the reconstructed scenes and design loss functions to promote depth estimation consistency across different views. Comprehensive experiments underscore our framework's superior generalization capabilities, surpassing existing state-of-the-art methods on several benchmark datasets without leveraging extra training information. Moreover, our innovative loss functions empower the model to autonomously recover domain-specific scale-and-shift coefficients using solely unlabeled images.
\end{abstract}

%%%%%%%%% BODY TEXT

%%%\vspace{-1.5em}
\section{Introduction}
%%\vspace{-0.5em} 
%background
Recovering 3D geometries of scenes from monocular images has become an area of significant interest, driven by recent advances in monocular depth estimation, with wide-ranging applications such as 3D photography~\cite{3dphoto}.
3D scene recovery~\cite{leres, yin2022reconstruction} of in-the-wild monocular  images relies on a powerful depth estimator~\cite{dpt, zamir2020robust, leres, ssi,zhang2022hierarchical} that can accurately predict the geometry of diverse scenes.
State-of-the-art depth estimation models~\cite{ssi, leres, dpt,zhang2022hierarchical} now advocate mix-dataset training~\cite{yin2020diversedepth, ssi}, which can generate robust depth predictions across diverse scenes. This opens up the possibility of large-scale pre-training and deploying only a single model in various application scenarios.

%SSI 
To enable mix-dataset training, scale-and-shift invariant (SSI) losses~\cite{ssi, leres} are designed to normalize depth representations explicitly, thereby removing scale-and-shift changes between different data sources. As a result, datasets with various depth representations, such as metric depth, uncalibrated disparity maps, and relative depth up to scale (UTS), can be jointly utilized for training.
However, despite the strong generalization capabilities across scenes, mix-dataset training comes with its own set of drawbacks. Unlike previous depth estimation models that produce absolute depth or relative depth up to scale, which can be directly unprojected to 3D structures given the intrinsic camera parameters, models trained with SSI losses predict depth up to unknown scale and shift factors (UTSS), which is geometrically incomplete~\cite{gp2} for reconstructing 3D models.
Although scaling depth maps typically adheres to the original 3D scene recovery's geometric integrity, the unknown shifts may introduce structural distortions.
Depth estimation models that are optimized to produce absolute depth or relative depth up to scale do not suffer from this problem, but they require geometry-complete depth annotations, such as metric depth or relative depth from multi-view stereo, for learning. Leres~\cite{leres} offers a potential solution by rectifying the distorted point cloud via a separately optimized post-processing module, which, however, necessitates additional 3D datasets.
Unfortunately, the extra 3D data or geometry-complete depth annotations are significantly less diverse than the geometrically incomplete data used in the original mix-data training, and as a result, their generalization ability on in-the-wild images is limited.

%our method 
In this research, our primary objective is to develop depth estimation models that can predict geometry-preserving depth up to a scale for 3D scene recovery without requiring extra data or annotations through mix-dataset training.
To achieve this goal, we propose a novel framework based on differentiable rendering. Specifically, we reconstruct 3D point clouds based on the predicted depth and use a differentiable renderer to generate novel views of the 3D model. We then predict the depth of the synthesized views with the same model and employ loss functions to ensure that the depth predictions of the rendered views are consistent. Fig.~\ref{fig:teaser} illustrates our motivation.
In this process, the network is optimized to produce undistorted 3D structures from depth using the informative gradients from the differentiable renderer. This ensures that the rendered images from different views look realistic and their depth estimations are consistent. Compared with previous works, our method can produce geometry-preserving predictions without relying on extra annotations or 3D datasets, enabling us to make full use of mixed datasets collected from various resources to improve generalization. Our loss functions can also recover domain-specific scale and shift coefficients of a trained UTSS model, such as Midas~\cite{ssi} and HDN~\cite{zhang2022hierarchical}, in a self-supervised manner using unlabelled images from the same domain.
Moreover, we demonstrate that our proposed self-supervised loss can be used to predict intrinsic camera parameters, such as focal length, by selecting the parameter from a few options that minimize the proposed consistency losses. Our extensive experiments on multiple benchmark datasets validate the effectiveness of our design. 
Our main contributions are summarized as follows:
 \begin{itemize}
 \itemsep -0.105cm
	\item We propose a novel depth estimation learning framework that can produce geometry-preserving depth without relying on extra datasets or annotations.
	\item  Our proposed consistency loss can recover domain-specific affine coefficients of a trained model in a self-supervised manner using unlabelled images from the same domain.
	\item We demonstrate that the self-supervised loss can also be used to roughly estimate camera intrinsic parameters.
    \item Experiments on multiple benchmark datasets show that our method can better recover scene structures of diverse images both quantitatively and qualitatively.
\end{itemize}

\begin{figure*}[h]
	\centering
	%left lower right upper.
	\includegraphics[trim=0cm 0cm 0cm 0cm, clip=true,width=1\linewidth]{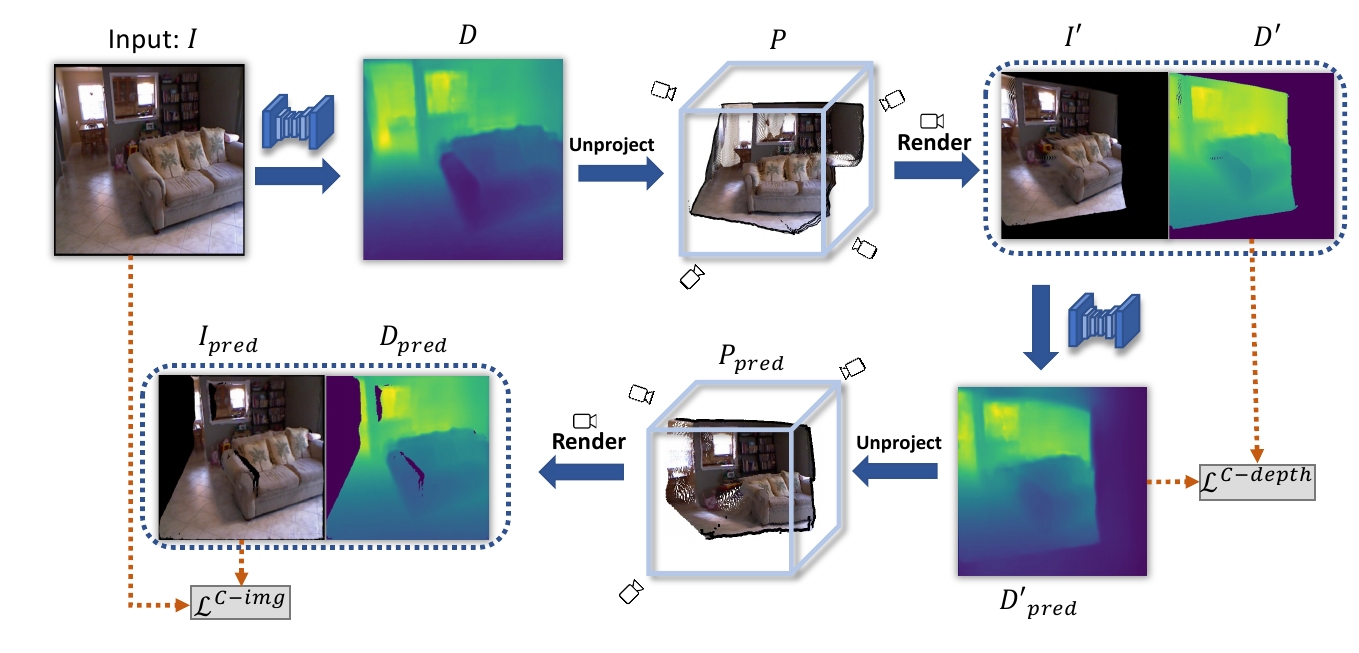}
% 	\vskip -1.5em
\vskip -1.5em
	\caption{\textbf{Overview of our framework for geometry-preserving depth estimation.} Given an input image, we reconstruct the point cloud based on the depth estimation of the image. Then, a new view of the 3D structure is rendered and the depth is estimated again using the same model. We then reconstruct the point cloud based on the depth estimation of the new view and render it back to the original view. Finally, loss functions are employed to promote consistency between the outputs from different views.}
\vskip -1em
	\label{fig:main}
\end{figure*}

%%\vspace{-0.8em} 

\section{Related Work}
%%\vspace{-0.5em} 
In this section, we review related works on monocular depth estimation based on deep neural networks and literature on 3D scene reconstruction from single-view images.

\textbf{Monocular depth estimation.} Deep neural networks have made significant progress in many computer vision tasks, including monocular depth estimation (MDE)~\cite{yuan2022new, Yin2019enforcing, li2022binsformer, bhat2021adabins, bian2019unsupervised, zamir2018taskonomy, eftekhar2021omnidata, DORN}, which is the focus of this paper. 
Most deep-learning-based  MDE models aim to learn a pixel-wise depth predictor by fitting a labeled training set.
However,  recent works\cite{ssi,leres, eftekhar2021omnidata, chen2020oasis, xian2018monocular} have shown that such models can be domain-sensitive and have poor generalization capability across datasets due to dataset bias.
Moreover, since collecting diverse labeled depth data at scale is difficult, the training datasets~\cite{chen2020oasis, zamir2018taskonomy, yin2020diversedepth, xian2020structure}  are often small, in sharp contrast to many other vision tasks, such as image segmentation~\cite{MSeg_2020_CVPR, li2022languagedriven, yin2022devil, OpenImages} and object detection~\cite{OpenImages, shao2019objects365}, which consistently benefit from the increasing number of diverse training data.
Recent studies~\cite{yin2021virtual, ssi, chen2020oasis, dpt, leres, zhang2022hierarchical} advocate mix-dataset training, which allows datasets from various sources to be jointly utilized for training. 
In particular, it is cheap to collect diverse stereo images at scale from the Internet~\cite{xian2018monocular} or 3D movies~\cite{ssi}. However, as the stereo cameras are not calibrated, the disparity map generated by 
optical flow algorithms can only recover inverse depths up to unknown scales and shifts.
Therefore, there exists a scale-and-shift gap between data annotations from various sources.

Self-supervised learning has been widely investigated in the area of depth estimation and 3D reconstruction~\cite{zhou2017unsupervised,godard2017unsupervised,mahjourian2018unsupervised}.
Compared with previous works, our focus is on generating robust 3D point clouds from monocular depth estimation through mix-data training, where we do not use additional information, such as video sequences or stereo images for self-supervision, and only a single monocular image is provided.

To enable mix-data training with various forms of annotations, such as metric depth~\cite{silberman2012indoor, geiger2012we, dai2017scannet} obtained from RGB-D cameras, disparity map from stereo matching~\cite{bauer2019uasol, cho2021diml}, and relative depth from multi-view stereo\cite{li2018megadepth},  state-of-the-art works~\cite{dpt, ssi,leres,zhang2022hierarchical} normalize the depth annotations explicitly, such that the scale-and-shift changes between depth annotations can be removed. In particular, HDN~\cite{zhang2022hierarchical} proposes to normalize depth annotations hierarchically to improve scale-and-shift invariant losses.
A significant advantage of mix-dataset training is that the learned model can be directly evaluated on various benchmarks without using their individual training sets, enabling zero-shot transfer~\cite{ssi}. 
However, the predicted depth is up to an unknown scale and shift, which needs to be aligned with the ground truth using least squares to find the optimal scale and shift, such that the standard evaluation metrics can be used for comparison. Alternatively, the zero-shot model can be finetuned with the training set of each benchmark for the evaluation of metric depth~\cite{ssi}.

\textbf{Recover 3D structure based on depth.}
Depth estimation is a crucial step in recovering the structure of a scene from monocular images. Our goal is to leverage the impressive generalization ability of zero-shot depth estimation models for 3D scene reconstruction in real-world settings.
While mix-dataset training has yielded good results in depth estimation, the learned model cannot be used directly for 3D scene reconstruction. This is because an unknown shift in depth can distort the structures.
 LeRes~\cite{yin2022reconstruction} addresses this problem by training a separate rectification module to correct the point cloud generated from raw depth maps. This module learns to shift the raw point cloud and predict the focal length given an initial value. However, this module requires an additional 3D point cloud dataset for training, which is hard to obtain for outdoor scenes and thus limits generalization across scenes.
GP2~\cite{gp2} demonstrates that using an extra scale-invariant loss on a portion of the dataset can help the model output geometry-preserving depth predictions that are automatically shifted. Adding a scale-invariant loss is the most straightforward and effective strategy to train a scale-invariant depth predictor. However, this loss requires extra geometry-complete depth annotations for supervision, which largely limits the model's generalization ability in diverse scenes. Consequently, the learned model is still domain-specific, which is contrary to the goal of learning a generic zero-shot model across scenes. Additionally, GP2~\cite{gp2} cannot estimate the focal length, which is essential for 3D reconstruction.

In contrast, our framework presented in this paper is complementary to previous mix-dataset training pipelines and can directly train a model that makes geometry-preserving depth predictions up to a scale without seeking extra datasets or annotations, which shows a strong generalization capability.

\textbf{Other 3D reconstruction in the literature.}
While our work focuses on 3D shape recovery from single images through depth estimation, there are various other methods that have been proposed to address this problem. Early works in this field relied mostly on hand-crafted features, such as local segments, shadings, and edges, as priors~\cite{karpenko2006smoothsketch, prados2005shape, saxena2008make3d}. More recent approaches, such as those proposed in~\cite{xiu2022icon, reizenstein2021common, yang2022banmo, saito2019pifu}, are data-driven and use an end-to-end learning approach. However, these methods are typically limited to specific types of scenes or objects, such as human bodies~\cite{saito2019pifu}, faces, cars, \etc, and require different forms of supervision, such as ground truth 3D structures represented by a mesh~\cite{wu2020unsupervised}.
\emph{One key advantage of our framework is that it is scene-agnostic, making it applicable to any real-world images.}

%%\vspace{-0.5em} 
\section{Preliminary}
%%\vspace{-0.5em} 
Before introducing our framework, we first present some preliminary concepts in monocular depth estimation and 3D scene reconstruction from depth.

\textbf{Scale-and-shift invariant loss.} To train a depth estimator that can be applied to various scenes, we utilize the scale-and-shift invariant loss~\cite{ssi} that eliminates the scale and shift variances between different data representations.
 Specifically, we first estimate the shift $\mu_{D}$ and scale $\sigma_{D}$ for a given depth map $D$ and a binary mask $M$ that indicates valid pixels, as follows:
 %%\vspace{-0.5em} 
\begin{equation}
\mu_{D}=\text{\tt median}(D), ~ ~\sigma_{D}=\frac{1}{|M|}\sum_{i=1}^{|M|}|D(i)-\mu_{D}|,
\end{equation}
where $D(i)$ is the depth value of a pixel location $i$, and $|M|$ is the number of valid pixels. We then remove the scale and shift for both the predicted depth map and the ground truth annotations, as shown below:
\begin{equation}
\hat{D}=\frac{{D}-\mu_{{D}}}{\sigma_{{D}}},~~\hat{D^{*}}=\frac{{D^{*}}-\mu_{{D^{*}}}}{\sigma_{{D^{*}}}}.
\end{equation}
Next, we apply a standard \Loss\ loss between the normalized depth representations to supervise the training, as follows:
\begin{equation}
    \mathcal{L}^{\text{SSI}}=|\hat{D} - \hat{D^{*}}|.
    \label{eq:ssi}
\end{equation}
This ensures that any affine changes to the predicted depth maps or the annotations will not affect the losses. Thus, uncalibrated disparity maps can be used for training.

\textbf{3D reconstruction from depth maps.} Given the predicted depth $D$, we can reconstruct the point cloud $P$ from the image coordinate based on the pinhole camera model, as shown below:
\begin{equation}
    \left\{\begin{matrix}x =&\hspace{-0.3cm} \frac{u-u_{0}}{f}d
    \\ y =&\hspace{-0.3cm} \frac{v-v_{0}}{f}d
    \\ z =&\hspace{-0.3cm} \hspace{0.75cm}d
    \end{matrix}\right.,
\label{eq: 3D point cloud reconstruction}
\end{equation}
where $(x,y,z)$ and $(u,v)$ are the 3D and 2D coordinates, $(u_{0}, v_{0})$ is the optical center, and $f$ is the focal length. As we can see, scaling the depth leads to uniform changes to the scene, while shifting causes non-uniform changes, which leads to distortion. Since the predicted depth maps are up to unknown scale and shift coefficients, the reconstructed 3D structure from them is likely to be distorted from inappropriate affine changes.

%%\vspace{-0.5em} 
\section{Method}
%%\vspace{-0.5em} 
This section presents a detailed description of our framework. Our framework comprises three crucial steps: depth estimation, point cloud reconstruction, and differentiable rendering of new views. Once we obtain a novel view of the structure, we repeat the same operations on the generated image and render it back to the original view. We then employ loss functions to promote consistency between the multi-view predictions. An overview of our framework is depicted in Fig.~\ref{fig:main}.

\subsection{Pipeline}

\textbf{Render a novel view of a point cloud.}
To obtain a novel view of a point cloud, we begin by reconstructing the point cloud $P$ using the predicted depth map $D$ of the input image $I$, as specified in Eq.~\eqref{eq: 3D point cloud reconstruction}. 
Next, we render a depth map $D^{\prime}$ and an RGB image $I^{\prime}$ of the point cloud from a different perspective.
Specifically, we rotate the camera horizontally by an angle $\theta$ and shift the camera along the $z$ axis by a distance $t$.
To ensure that the scene remains in the rendered image, we rotate the camera around $T_{\text{\tt center}}=[0, 0, {\min}_z (P)]^{\text{T}}$ instead of rotating around the origin of coordinates. Here, ${\min}_z (P))$ returns the smallest depth value along the $z$ axis.
In this way, the structure always appears in the middle of the rendered image during rotation.
More specifically, the rotation matrix $R$ and transition matrix $T$ are given below:
\begin{equation}
    R=\begin{bmatrix}
{\cos} (\theta) & 0 & {\sin} (\theta)\\
0  & 1 & 0\\
 -{\sin} (\theta) & 0 & {\cos} (\theta)\\
\end{bmatrix},  \,\,
T=\begin{bmatrix}
0\\
 0\\
t\\
\end{bmatrix}.
\end{equation}
Here, the shift $t$ and rotation angle $\theta$ are randomly sampled from $[-{\min}_z (P),2 \cdot {\min}_z (P)]$ and $[-30^{\circ}, 30^{\circ}]$, respectively.
Finally, we use the differentiable renderer~\cite{wiles2020synsin}, which is a function of the point cloud data, the rotation matrix, and the transition matrix, to render the new image $I^{\prime}$ and the depth map $D^{\prime}$ as follows:
\begin{equation}
I^{\prime}, D^{\prime} = \text{\tt Render} (P-T_{\text{\tt center}}, R, T+T_{\text{\tt center}}),
\end{equation}
To implement the change of the rotation center conveniently, we shift the point cloud data using $P-T_{\text{\tt center}}$ and compensate for the shift by adding it to the shift matrix after rotation: $T+T_{\text{\tt center}}$.

\textbf{Render the raw view.}
Given the rendered image $I^{\prime}$ and the depth map $D^{\prime}$ of new views, we can recover the same  structure $P$ and accurately render back the raw view, except for occluded regions.
Based on this observation,  we estimate the depth of the rendered image $I^{\prime}$  with the same model to obtain $D^{\prime}_{\text{\tt pred}}$, and use it together with $I^{\prime}$ to render back the raw view. 
Our intuition is that we can achieve accurate recovery of the structure and rendering of the raw view only if the predicted depth $D^{\prime}_{\text{\tt pred}}$ of the rendered image matches the rendered depth map $D^{\prime}$.
To achieve this goal,  both the reconstructed structure $P$ and the rendered image $I^{\prime}$ must be geometrically accurate and visually realistic.
Therefore, we render back the raw view based on $I^{\prime}$ and $D^{\prime}_{\text{\tt pred}}$.
However, due to the scale-and-shift changes of the depth predictions output by the model trained with SSI loss, we need to align $D^{\prime}_{\text{\tt pred}}$ with $D^{\prime}$ using the least squares method to ensure consistency:
\begin{align}
  (a, b) &= \arg \min_{a,b} \sum_{i=1}^{|M^{\prime}|} (aD^{\prime}_{\text{\tt pred}}(i)+b-D^{\prime}(i) ),\\
  \hat{D}^{\prime}_{\text{\tt pred}}&=aD^{\prime}_{\text{\tt pred}}+b.
\end{align}
Specifically, we solve for $a$ and $b$ in the equation above, where $|M^{\prime}|$ is the number of valid pixel locations.
We can  then reconstruct the structure again based on  $I^{\prime}$ and $\hat{D}^{\prime}_{\text{\tt pred}}$ to obtain $P_{\text{\tt pred}}$. 
Finally, we render the depth map $D_{\text{\tt pred}}$ and the image $I_{\text{\tt pred}}$ of the raw view based on the point cloud $P_{\text{\tt pred}}$:
\begin{align}
    I_{\text{\tt pred}}, D_{\text{\tt pred}} &= \text{\tt Render} (P_{\text{\tt pred}},R_{\text{\tt inv}},T_{\text{\tt inv}}),\\
    R_{\text{\tt inv}}&=R^{\text{T}},\\
    T_{\text{\tt inv}}&=-R^{\text{T}}(T+T_{\text{\tt center}})+T_{\text{\tt center}}.
\end{align}

\subsection{Loss Functions}
%%\vspace{-0.5em} 
\textbf{Consistency Loss.}
To ensure consistency during the rendering process, we employ two loss functions: the depth consistency loss $\mathcal{L}^{\text{C-depth}}$ and the image consistency loss $\mathcal{L}^{\text{C-img}}$. The overall consistency loss is defined as follows:
\begin{equation}
\mathcal{L}^C=\mathcal{L}^{\text{C-depth}}+\alpha \mathcal{L}^{\text{C-img}},
\end{equation}
where $\alpha$ is a weight term.
For the image consistency loss, we compute the mean pixel-wise \Loss\  loss ($\text{\tt L1}$) between the rendered image $I_{\text{\tt pred}}$ and the original input image $I$ over the valid region $M_{\text{img}}$:
   $ \mathcal{L}^{\text{C-img}}={\tt L1 }_{M_{\text{img}} } (I_{\text{\tt pred}}, I)$.
For the depth consistency loss, we apply a scale-and-shift invariant loss ($\text{\tt SSI}$) in Eq.~\ref{eq:ssi} between  the rendered depth map $D^{\prime}$ of the novel view and the new prediction $D^{\prime}_{\text{\tt pred}}$ over the valid regions $M_{\text{depth}}$: $\mathcal{L}^{\text{C-depth}}=\text{\tt SSI}_{M_{\text{depth}}} ( D^{\prime}, D^{\prime}_{\text{\tt pred}}  )$.
Here, $M_{\text{img}}$ and $M_{\text{depth}}$ include the pixels that are not occluded during rendering.

\textbf{Multi-Focal-Length (MFL) losses.}
\label{sec:multifov}
Our pipeline and loss computation rely on knowing the focal length $f$ of the input image to unproject the depth map.  Although the intrinsic camera parameters are provided in some datasets, they are often unknown in many cases, such as web images.
Availability of focal length can definitely make our design more robust but it meanwhile limits the training in the mix-dataset scenario. 
Therefore, we assume that the focal length of training images is unknown during training, which allows us to train our algorithm with more diverse data.
To handle the unknown and varied focal length of training images, we further develop a multi-focal-length loss  for our algorithm.
A focal length $f$ can be roughly estimated from the horizontal field of view ($\rm FOV$), such as $60^\circ$, by
\begin{equation}
    f=\frac{W}{2\cdot{\tan}({\rm FOV}/2)},
    \label{eq:fov}
\end{equation}
where $W$ is the width of the image. Our proposed consistency loss based on the  focal length $f$ is denoted as $\mathcal{L}^C_{f}$, and we can compute multiple consistency losses by selecting different foal length values from a set $\mathcal{F}$ that corresponds to different FOVs.
At training time, we only keep the minimum one as the final loss of each input image $\mathcal{L}^C_{f^{*}}$ for optimization, where
\begin{align}
  f^{*} &= \arg \min_{f\in\mathcal{F}} \mathcal{L}^C_{f}.
\end{align}
Intuitively, the appropriate focal length can more accurately reconstruct the structure  and thus leads to small rendering losses. 
Finally, we add the raw scale-and-shift invariant loss between the original predicted depth map $D$ and its ground truth $D^{*}$, and the final loss is given by:
\begin{equation}
    \mathcal{L}= \mathcal{L}^{\text{SSI}}+\beta\mathcal{L}^C_{f^{*}},
\end{equation}
where $\beta$ is a hyper-parameter to balance  two terms.

\begin{table*}[t]
\centering
% \resizebox{\linewidth}{!}{%
\begin{tabular}{ l |c|ll|ll|ll|ll}
\toprule[1pt]
\multirow{2}{*}{Method} & \multirow{2}{*}{Extra Info.} & \multicolumn{2}{c|}{NYU} & \multicolumn{2}{c|}{ScanNet} & \multicolumn{2}{c|}{ETH3D} & \multicolumn{2}{c}{KITTI}   \\
 &   & AbsRel$\downarrow$     & $\delta_{1}$$\uparrow$     & AbsRel$\downarrow$      & $\delta_{1}$$\uparrow$      & AbsRel$\downarrow$      & $\delta_{1}$$\uparrow$      &AbsRel$\downarrow$      & $\delta_{1}$$\uparrow$       \\ 
 \toprule[1pt]
 SSI  & -  & $27.9$ & $52.5$ & $28.5$ & $51.8$ &  $26.3$ & $62.5$ & $21.0$ & $63.9$ \\ 
 SSI +  PCM~\cite{leres} & 3D dataset  & $13.5$ & $83.4$ & $12.4$ & $85.2$ &  $20.6$ & $71.2$ & $33.7$ & $39.2$ \\ 
 \textbf{SSI + Ours} & - & $\textbf{10.5}$ & $\textbf{89.2}$ & $\textbf{12.2}$ & $\textbf{85.4}$ &  $\textbf{11.3}$ & $\textbf{87.1}$ & $\textbf{12.9}$ & $\textbf{81.4}$ \\  \hline \hline
GP2~\cite{gp2} & Metric Depth  & $9.8$ & $90.3$ & $11.4$ & $87.2$ &  $14.0$ & $85.4$ & $13.4$ & $81.2$ \\ 
  \textbf{GP2 + Ours} & Metric Depth  & $\textbf{9.2}$ & $\textbf{91.2}$ & $\textbf{10.9}$ & $\textbf{88.0}$ &  $\textbf{10.9}$ & $\textbf{88.4}$ & $\textbf{11.6}$ & $\textbf{84.9}$ \\ 
 \toprule[1pt]
\end{tabular}
% }
\vspace{-0.7em}
\caption{
\textbf{Comparison of different methods for geometry-preserving depth estimation with scale alignment only.} Our method outperforms previous methods without using any additional information. 
}
\label{tab:depth}
\vspace{-1em}
\end{table*}

\begin{table}[t]
\centering
\resizebox{0.40\textwidth}{!}{%
\begin{tabular}{l|lll}
\toprule[1pt]
\multicolumn{1}{c|}{\multirow{2}{*}{Method}}  & KITTI  & 2D3D &NYU  \\
\multicolumn{1}{c|}{}                        & \multicolumn{3}{c}{RMSE$\downarrow$}           \\ 
\toprule[1pt]
PCM~\cite{leres}   & 7.05    & 0.67 &0.38  \\
\textbf{RenderFOV \textit{w/} $\mathcal{L}^{\text{C-img}}$ }  & 5.95 &	0.72 & 0.43 \\
\textbf{RenderFOV \textit{w/} $\mathcal{L}^{\text{C-depth}}$ } &  \textbf{3.04} &	\textbf{0.48}  & \textbf{0.36} \\
\toprule[1pt]
\end{tabular}
}
\vspace{-0.7em}
\caption{
Comparison of different methods for point cloud reconstruction. Our method with $\mathcal{L}^{\text{C-depth}}$ for selecting the FOV has the optimal results on indoor and outdoor benchmarks.
}
\label{tab:3d}
\end{table}

\begin{table}[t]
\centering
\resizebox{0.30\textwidth}{!}{%
\begin{tabular}{cccc}
\toprule[1pt]
 $\mathcal{L}^{\text{C-img}}$ & $\mathcal{L}^{\text{C-depth}}$ & MFL & AbsRel$\downarrow$ \\ 
 \toprule[1pt]
 \checkmark      & \checkmark      & \checkmark      &    $11.7$ \\
 \checkmark      & \checkmark      &       &    $13.3$ \\
\checkmark     &        &    \checkmark     &    $13.4$     \\
&   \checkmark     &      \checkmark  &    $12.2$     \\  
&        &        &    $26.0$     \\  
\toprule[1pt]
\end{tabular}%
}
\vspace{-0.7em}
\caption{\textbf{Ablation study on loss functions.} We report the mean AbsRel over four benchmarks for evaluation.  Our designs can provide remarkable performance improvement.
}
\label{tab:ablation}
\end{table}

\begin{table}[t]
\centering
\resizebox{0.30\textwidth}{!}{%
\begin{tabular}{lr|ll}
\toprule[1pt]
\multicolumn{1}{c}{\multirow{2}{*}{Method}} &\multicolumn{1}{c|}{\multirow{2}{*}{Eval}}  & NYU  &   KITTI   \\
\multicolumn{1}{c}{}    &\multicolumn{1}{c|}{}                        & \multicolumn{2}{c}{AbsRel$\downarrow$}           \\ 
\toprule[1pt]
SSI  & SSA & 9.3  & 14.0     \\\hline
 SSI  & SA & 27.9  & 21.0     \\
\textbf{SSL \textit{w/} $\mathcal{L}^{\text{C-img}}$ } & SA & 10.8 & 13.3   \\
\textbf{SSL \textit{w/} $\mathcal{L}^{\text{C-depth}}$ } & SA & 10.8  & 13.1     \\
\toprule[1pt]
\end{tabular}
}
\vspace{-0.7em}
\caption{
\textbf{Self-supervised learning (SSL) experiments.} We only learn the affine coefficients to scale and shift the prediction of a model trained by the SSI loss  with the unlabelled training images in individual benchmarks. \textbf{SSA} and  \textbf{SA} denote scale-and-shift alignment and scale alignment, respectively, for evaluation.
}
\label{tab:self}
\end{table}

% Quantitative comparison of depth with SOTA methods
\begin{table*}[t]
\setlength{\tabcolsep}{2pt}
\small
\centering
% \resizebox{0.8\linewidth}{!}{%
\begin{tabular}{ r |ll|ll|ll|ll}
\toprule[1pt]
\multirow{2}{*}{Method}  & \multicolumn{2}{c|}{NYU} & \multicolumn{2}{c|}{KITTI} & \multicolumn{2}{c|}{ScanNet} & \multicolumn{2}{c}{ETH3D}  \\
   & AbsRel$\downarrow$     & $\delta_{1}$$\uparrow$     & AbsRel$\downarrow$      & $\delta_{1}$$\uparrow$      &AbsRel$\downarrow$      & $\delta_{1}$$\uparrow$       &AbsRel$\downarrow$     & $\delta_{1}$$\uparrow$   \\ \hline
MegaDepth~\cite{li2018megadepth}&$19.4$& $71.4$ &$20.1$ &$66.3$  &$26.0$  &$64.3$ &$39.8$ &$52.7$\\
WSVD~\cite{wang2019web}  &$22.6$ &$65.0$ &$24.4$ &$60.2$  &$18.9$ &$71.4$ &$26.1$ &$61.9$ \\
DiverseDepth~\cite{yin2021virtual}  &$11.7$ &$87.5$ &$19.0$ &$70.4$ &$10.8$ &$88.2$ &$22.8$ &$69.4$\\
MiDaS~\cite{ssi}  &$11.1$ &$88.5$ &$23.6$ &$63.0$ &$11.1$ &$88.6$  & $18.4$ &$75.2$ \\
Leres~\cite{leres}    &${9.0}$  &${91.6}$  &${14.9}$ &${78.4}$ &${9.5}$ &${91.2}$ &${17.1}$ &${77.7}$ \\
DPT$^\dagger$~\cite{dpt}    &${8.8}$  &${92.7}$  &${12.7}$ &${84.9}$ &${9.6}$ &${91.3}$ &${16.1}$ &${77.6}$ \\
\hline
\textbf{Ours (SA)}   & $\textbf{7.7}$  & $\textbf{93.6}$  & $\textbf{11.1}$ & $\textbf{86.2}$&$\textbf{9.0}$ &$\textbf{91.8}$ &$\textbf{9.1}$ &$\textbf{91.4}$ \\
\hline
 \toprule[1pt]
\end{tabular}
% }
%\vspace{-1em}
\caption{
\textbf{Comparison with the state-of-the-art depth estimation models on zero-shot benchmarks. }
Our geometry-preserving model with only scale alignment (\textbf{SA}) significantly outperforms previous works with scale-and-shift alignment for evaluation. $^\dagger$ denotes our re-implementation with our data.
}
\label{tan:sot}
\vspace{-1.5em}
\end{table*}

%%\vspace{-0.5em} 

\section{Experiments}
%%\vspace{-0.5em} 
We conduct extensive experiments on multiple benchmarks to validate the effectiveness of our algorithms, including evaluations of the geometry-preserving depth estimation and the accuracy of the recovered point clouds. Additionally, we conduct several ablation studies to analyze each component of our design.

\textbf{Implementation details.}
We follow Leres~\cite{leres} and GP2~\cite{gp2} to construct datasets for mix-dataset training, which contain 121K images from DIML~\cite{kim2018deep} dataset, 114K images from taskonomy~\cite{zamir2018taskonomy} dataset, 48K images from Holopix50K~\cite{hua2020holopix50k}, and 20K images from HRWSI~\cite{xian2020structure} dataset. We withhold 200 images from each dataset for validation. For ablation study and analysis, we sample a subset of 16,000 images evenly from different datasets to train the models.
All baselines in our experiments and our model employ the DPT~\cite{dpt} depth estimation network, which is currently state-of-the-art, with an input image size of $384\times384$ pixels. During training, we apply random cropping and horizontal flipping for data augmentation.
Our training process consists of two stages: first, we fully optimize our model for 20 epochs using the SSI loss, as in previous mix-dataset training methods. Second, we finetune the model by adding our proposed losses for 2 additional epochs. 
Following DPT~\cite{dpt}, we use the Adam optimizer~\cite{kingma2014adam}, with a learning rate of $10^{-5}$ and the batch size of 16.
We construct mini-batches by sampling equal numbers of data from each dataset.
The weight terms $\alpha$ and $\beta$ are set to 1.0 and 0.1, respectively. We select three different fields of view (FOV) from $\{50^\circ, 60^\circ, 70^\circ\}$ during training and compute their corresponding focal length values based on Eq.~\ref{eq:fov} to construct the focal length set $\mathcal{F}$.

\textbf{Evaluation Metrics.}
We evaluate our model on five benchmark datasets, including NYU V2~\cite{silberman2012indoor}, ScanNet~\cite{dai2017scannet}, KITTI~\cite{geiger2012we}, ETH3D~\cite{schops2017multi}, and 2D3D~\cite{armeni2017joint}. For evaluation of geometry-preserving depth estimation, we follow Leres~\cite{leres} to first align the scale of the predicted depth map $D$ and the ground truth $D^*$ by multiplying the prediction with a factor $s$, which is computed by: 
\[
s= \text{\tt median} ({D^*}/{D}). 
\]
We use two common metrics to evaluate the accuracy: the absolute relative error (AbsRel): 
\[ (1/|M|) \sum_{i=1}^{|M|} |D(i)-D(i)^*|/D(i)^*. \]
and the percentage of pixels with 
\[ 
\delta_{1}={\max}({D(i)}/{D^{*}(i)}, {D^{*}(i)}/{D(i)})<1.25.
\] 
For evaluation of the reconstructed point clouds, we compute the Root Mean Square Error (RMSE) between the point clouds unprojected from the aligned depth prediction and the ground-truth following ~\cite{gp2}. 

\subsection{Results}
%%\vspace{-0.5em} 
\textbf{Evaluation of geometry-preserving depth estimation. } To demonstrate the advantages of our design, we compare our model with several existing methods, including:

1) \textbf{SSI.} Our first baseline is the model trained with the standard SSI loss. As the output from it is up to unknown scale and shift coefficients, we can observe the influence when the shift is omitted.  

2) \textbf{SSI + PCM.} We incorporate the point cloud module (PCM) from Leres~\cite{leres} into our SSI baseline to predict the shifts required to rectify the distorted point clouds generated by depth estimation models. In this analysis, we used their released PCM model to shift the depth maps generated by our SSI baseline.

3) \textbf{GP2}. 
Although our learning approach focuses on developing geometry-preserving depth estimators without relying on geometry-complete depth annotations, we also compared our method to GP2, which directly utilizes scale-invariant (SI) losses, requiring geometry-complete depth annotations. Since the metric depth annotations were only available in the Taskonomy dataset, we applied the SI loss to the data from this dataset in each mini-batch during training. The comparison of the results is presented in Table~\ref{tab:depth}.

The results indicate that directly omitting the shift term of the SSI baseline resulted in poor performance across all benchmarks, implying that the model trained with SSI alone cannot be used to reconstruct 3D structures directly.
Our proposed design outperforms PCM~\cite{leres} across all benchmarks, especially in those with outdoor images such as KITTI and ETH3D. This is because PCM is trained with 3D indoor datasets and is only effective in indoor datasets, such as NYU and ScanNet.
Utilizing scale-invariant losses with metric depth labels can effectively improve the performance on indoor benchmarks, but is less effective than our model on outdoor benchmarks. In particular, our result in terms of AbsRel outperforms GP2 by $19.2\%$ on the ETH3D dataset. Our design can further improve the performance when they are combined, which generates optimal results across all benchmarks.

\textbf{Evaluation of reconstructed point clouds.} We next evaluate the accuracy of the reconstructed point clouds. Given the predicted depth map, the focal length of the image needs to be estimated for point cloud reconstruction. Here we use the same  depth estimation model learned by our losses and compare the following strategies to generate the focal length values.

1) \textbf{PCM} from Leres~\cite{leres}. In Leres~\cite{leres}, they reconstruct the point clouds using an initial focal length value, which corresponds to a FOV of $60^{\circ}$. Then, the PCM takes the point cloud as input and predicts shifts of the points, as well as focal length values. These values are used to reconstruct the point cloud again for evaluation.

2) \textbf{RenderFOV.} 
We use the proposed consistency losses as an indicator to choose the focal length values from a few selections during testing.
Here we provide multiple FOV values from $\{40^{\circ},50^{\circ},60^{\circ},70^{\circ},80^{\circ}\}$, and compute their corresponding focal length values. 
We render multiple images to compute the mean consistency losses, by
 selecting the camera shifts $t$ from $\{-0.5 \cdot {\min}_z (P),0.5 \cdot {\min}_z (P)\}$ and rotation angles $\theta$ from  $\{-20^{\circ},20^{\circ}\}$.
 We choose the focal length value that produces minimal  consistency loss. We report results using the image consistency loss and depth consistency losses, respectively.

Table~\ref{tab:3d} shows the comparison of the  methods. We find that the depth consistency loss is more effective than the image consistency loss for selecting the appropriate focal length values. Therefore, using depth consistency loss becomes our default choice for point cloud reconstruction.
Our proposed method outperforms the previous method PCM~\cite{leres}, particularly in the outdoor dataset KITTI.

\begin{figure*}[t]
	\centering
	%left lower right upper.
	\includegraphics[trim=0cm 0cm 0cm 0cm, clip=true,width=0.98\linewidth]{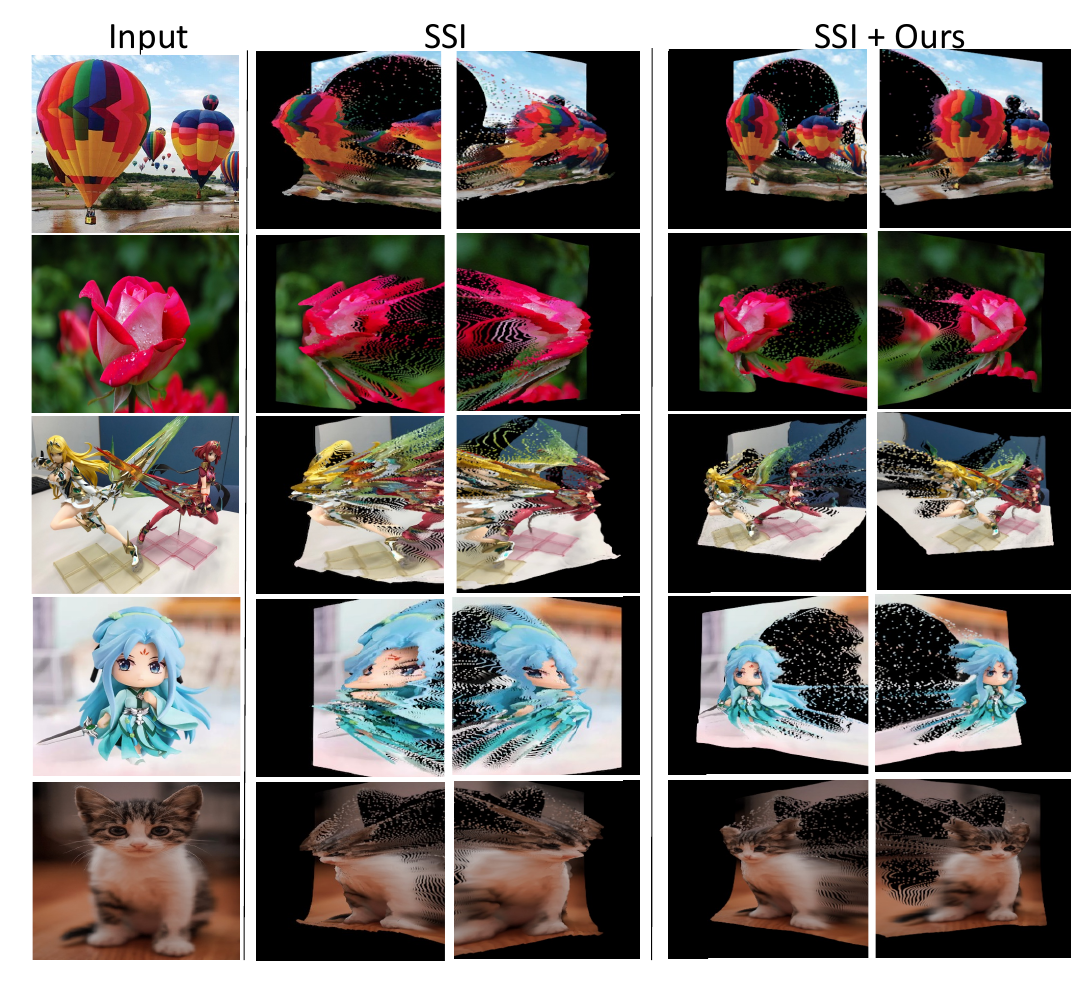}
	%\vskip -0.8em
	\caption{\textbf{Qualitative comparison of point cloud reconstruction.  } We render a new view of the reconstructed point clouds. Adding our loss in training can effectively improve the baseline. 
 }
	\label{fig:visulize}
%%\vspace{-1.5em}
\end{figure*}

\textbf{Self-supervised learning for scale-and-shift recovery.}
We explore self-supervised learning  with our framework using unlabelled images. Here we fix the parameters of a fully trained model with SSI loss, and learn the affine parameters to transform the raw outputs up to scale and shift coefficients to geometry-preserving outputs up to a scale.
 We observe the optimal scales and shifts of different benchmarks are different, hence we learn a group of affine parameters for each benchmark based on their individual training sets. Specifically, we conduct experiments on two datasets: NYU, an indoor dataset containing 795 training images, and KITTI, an outdoor dataset containing 21,591 training images.
We also report the results of the pre-trained model with scale-and-shift alignment for evaluation, which can be considered as the upper bound for the results of our models with only scale alignment. The results are presented in Table~\ref{tab:self}.

Our self-supervised learning with both losses yields good results on both benchmarks, indicating the effectiveness of our proposed losses and suggesting that predictions of images from the same domain have similar affine coefficients.
Notably, our self-supervised learning models on the KITTI dataset even outperform the SSI pre-trained model with scale-and-shift alignment. This result suggests that using least squares to find the affine coefficients is less effective than our approach in this case.
One possible explanation is that least-squares regression requires accurate predictions for each image to obtain the affine parameters for alignment. When the raw prediction is poor, the errors may be amplified by alignment, resulting in sub-optimal results. In contrast, optimizing the affine parameters with our losses based on the images from the same domain yields better results.

\textbf{Qualitative results.} We next compare the 3D structures reconstructed by the model trained with SSI loss and our losses through visualization in Fig.~\ref{fig:visulize}. Specifically, we show the rendered images of novel views. Our proposed losses can effectively eliminate distortions in the 3D models, as shown in the visualization.

\textbf{Comparison with the state-of-the-art methods.}
We compare our best model with state-of-the-art methods on four benchmark datasets: NYU, KITTI, ScanNet, and ETH3D. We use all the training data in our mixed datasets and add the scale-invariant loss~\cite{gp2} during training. As the training data of DPT~\cite{dpt} are not publicly available, we re-implement DPT with our datasets and training hyper-parameters. Previous methods align the scale and the shift for evaluation, while our methods only require alignment of the scale.
As shown in Table~\ref{tan:sot}, our method outperforms the state-of-the-art with significant performance advantages. Notably, our performance on the challenging ETH3D dataset surpasses the second-best result by $47\%$ (AbsRel). Our model demonstrates strong generalization ability in various benchmarks while also providing geometry-preserving predictions, which show practical values.
%%%\vspace{-0.5em} 

\section{Conclusion}
%%%\vspace{-0.5em} 
In this paper, we propose a new geometry-preserving depth estimation framework that supports large-scale mix-dataset training without requiring extra information. The proposed multi-view consistency loss can recover the domain-specific affine parameters of a trained model in a self-supervised manner. It can also roughly estimate the focal length by selecting appropriate values from a few selections. Our experiments on multiple benchmarks validate the effectiveness of our design.

\section*{Acknowledgements}
Part of this work was supported by the National Key R\&D Program of China (No.\ 2022ZD0118700).

{\small
\bibliographystyle{ieee_fullname}
\bibliography{egbib}
}

\end{document}